\pdfoutput=1

\documentclass[11pt]{article}

\usepackage[final]{acl}
\usepackage{amsfonts}
\usepackage{times}
\usepackage{latexsym}
\usepackage{inconsolata}
\usepackage{amsmath}
\usepackage{graphicx}
\usepackage{comment}
\usepackage{amssymb}
\usepackage[inkscapeformat=png]{svg}
\usepackage{paralist}
\usepackage{svg}
\usepackage{enumitem}
\usepackage{listings}
\usepackage{multicol}
\usepackage{booktabs}
\usepackage{multirow}

\usepackage[T1]{fontenc}

\usepackage[utf8]{inputenc}

\usepackage{microtype}

\usepackage{inconsolata}

\usepackage{graphicx}

\usepackage{listings}
\usepackage{xcolor}

\definecolor{codegreen}{rgb}{0,0.6,0}
\definecolor{codegray}{rgb}{0.5,0.5,0.5}
\definecolor{codepurple}{rgb}{0.58,0,0.82}
\definecolor{backcolour}{rgb}{0.95,0.95,0.92}

\lstdefinestyle{mystyle}{
    backgroundcolor=\color{backcolour},   
    commentstyle=\color{codegreen},
    keywordstyle=\color{magenta},
    numberstyle=\tiny\color{codegray},
    stringstyle=\color{codepurple},
    basicstyle=\ttfamily\tiny,
    breakatwhitespace=false,         
    breaklines=true,                 
    captionpos=b,                    
    keepspaces=true,                 
    numbers=right,                   
    numbersep=5pt,                  
    showspaces=false,                
    showstringspaces=false,
    showtabs=false,                  
    tabsize=1
}

\newtheorem{theorem}{Theorem}[section]

\newtheorem{lemma}[theorem]{Lemma}
\newenvironment{proof}{{\bf Proof.}}{\hfill\rule{2mm}{2mm}}

\title{Offline Preference Optimization \\ via Maximum Marginal Likelihood Estimation}

\author{Saeed Najafi \and
  Alona Fyshe \\
  Department of Computing Science, University of Alberta, Canada\\
  \texttt{\{snajafi,alona\}@ualberta.ca} \\}

\begin{document}
\maketitle

\begin{abstract}
Aligning Large Language Models (LLMs) with human preferences is crucial, but standard methods like Reinforcement Learning from Human Feedback (RLHF) are often complex and unstable. In this work, we propose a new, simpler approach that recasts alignment through the lens of Maximum Marginal Likelihood (MML) estimation. Our new \textbf{MML-based Preference Optimization (MMPO)} maximizes the marginal log-likelihood of a preferred text output, using the preference pair as samples for approximation, and forgoes the need for both an explicit reward model and entropy maximization. We theoretically demonstrate that MMPO implicitly performs preference optimization, producing a weighted gradient that naturally up-weights chosen responses over rejected ones. Across models ranging from 135M to 8B parameters, we empirically show that MMPO: 1) is more stable with respect to the hyperparameter $\beta$ compared to alternative baselines, and 2) achieves competitive or superior preference alignment while better preserving the base model’s general language capabilities. Through a series of ablation experiments, we show that this improved performance is indeed attributable to MMPO's implicit preference optimization within the gradient updates.
\end{abstract}

\section{Introduction}
LLMs have demonstrated remarkable capabilities across a wide range of tasks. This success is largely due to a multi-stage training process involving pre-training, supervised fine-tuning, and preference alignment~\cite{ouyang2022training}. The final alignment stage, which ensures that model outputs are helpful, harmless, and consistent with human preferences, is critical for real-world deployment. The predominant paradigm for this alignment has been RLHF~\cite{ouyang2022training}, typically implemented with complex policy gradient algorithms like Proximal Policy Optimization (PPO)~\cite{schulman2017proximalpolicyoptimizationalgorithms}. However, the RLHF pipeline is notoriously challenging; it requires training a separate reward model, which can be unstable and poorly calibrated, and its PPO-based optimization is sensitive to hyperparameters and computationally expensive~\cite{engstrom2020implementationmattersdeeppolicy, patterson2024empiricaldesignreinforcementlearning}.

Recent advancements have sought to simplify this pipeline. Direct Preference Optimization (DPO) marked a significant breakthrough by reframing preference learning as a direct classification problem on human preference data~\cite{rafailov2023direct}. This approach elegantly sidesteps the need for an explicit reward model and the instabilities of reinforcement learning, demonstrating strong empirical performance. The success of DPO motivates a natural question: can the alignment objective be simplified even further? While DPO removes the reward model, its loss is still derived directly from the RLHF objective, which implicitly encourages optimizing the entropy of the language model. We investigate whether effective alignment can be achieved under even more relaxed assumptions, removing the explicit need for both a reward function and the entropy maximization term inherent in the standard RLHF objective.

In this paper, we propose \textbf{MMPO}, a novel and simple algorithm for offline preference alignment. Drawing inspiration from the principle of MML estimation~\cite{guu-etal-2017-language}, MMPO recasts the alignment problem. Instead of directly optimizing for preferences, we maximize the marginal log-likelihood of a preferred text output, where the given preference pair is treated as two samples to approximate this marginalization. We show theoretically (Theorem~\ref{theory:implicit_po_mml}) that this MML objective, when optimized with a numerically stable `log-sum-exp' formulation, implicitly performs preference optimization. The resulting gradient updates naturally up-weight the preferred response higher than the dispreferred one, modulated by a sigmoid of their score difference, without explicitly formulating a reward or an entropy-maximizing term.

Our extensive experiments on models ranging from 135M to 8B parameters demonstrate the effectiveness and stability of MMPO compared to strong baselines like DPO and SimPO~\cite{meng2024simpo}. Our main contributions are as follows:
\begin{enumerate}[label=\arabic*), topsep=0pt, itemsep=0pt, parsep=0pt]
    \item We introduce \textbf{MMPO}, a new offline preference optimization algorithm, and provide a theoretical analysis showing that the MMPO on preference pairs is equivalent to a weighted gradient update that inherently favors preferred responses.
    \item We empirically show that MMPO is \textbf{more stable} when varying the key hyperparameter $\beta$ (compared to both DPO and SimPO).
    \item We find that MMPO \textbf{better preserves the general language capabilities} of the base instruction-tuned models.
    \item We demonstrate that MMPO achieves \textbf{strong and consistent preference alignment}, reaching competitive or superior win-rates on the AlpacaEval-2 benchmark.
\end{enumerate}

This paper is structured as follows. We first review the preliminaries of RLHF in Section \ref{sec:methods}. In Section \ref{sec:max_marginal_likelihood}, we introduce the MML framework, present our core theoretical result showing its connection to preference optimization, and derive our proposed MMPO algorithm. Finally, in Section~\ref{experiments}, we empirically validate MMPO, demonstrating its superior stability and its effectiveness at balancing alignment with reasoning and common-sense capability retention.

\section{Methods}
\label{sec:methods}

In our learning setting, we are provided with a dataset of input-output text pairs $(x, y)$, where $x$ represents an input text and $y$ denotes its corresponding target text completion. We assume the existence of an oracle reward function $r(y, z)$, which quantifies the alignment between a generated sample $z$ and the target completion $y$. The objective for RLHF \cite{ouyang2022training} is formulated as a constrained optimization problem. In this problem, a language model $\pi_\theta$, parameterized by $\theta$, is updated by maximizing the following objective function $J_{\text{RLHF}}(\theta) = $
\begin{multline}
\label{eq:rlhf_objective}
\mathbb{E}_{z \sim \pi_\theta(\cdot|x)} \left [ r(y, z) \right ] - \beta \mathbb{D}_{KL} \left [ \pi_\theta(\cdot|x) || \pi_{\text{sft}}(\cdot|x) \right ] \\
= \mathbb{E}_{z \sim \pi_\theta(\cdot|x)} \left [ r(y, z) - \beta \log \frac{\pi_\theta(z|x)}{\pi_{\text{sft}}(z|x)} \right ]
\end{multline}

In the RLHF objective, $\mathbb{E}_{z \sim \pi_\theta(\cdot|x)}$ denotes the expectation over text completions $z$ sampled from the language model $\pi_\theta$ conditioned on input $x$.  The term $\mathbb{D}_{KL}$ represents the Kullback-Leibler divergence, measuring the difference between the policy $\pi_\theta$ and $\pi_{\text{sft}}$, a Supervised Fine-Tuned (SFT) language model that serves as the initial checkpoint for $\pi_\theta$. By maximizing the objective function $J_{\text{RLHF}}(\theta)$, we update the main model $\pi_\theta$ to generate text completions $z$ that achieve higher expected rewards while remaining close to the initially trained model $\pi_{\text{sft}}$. The hyper-parameter $\beta \ge 0$ introduces the KL constraint to balance between reward maximization and minimizing the deviation from $\pi_{\text{sft}}$. For further background discussion on RLHF and DPO, please see Appendix~\ref{appendix:methods_dpo}.

\section{Maximum Marginal Likelihood}
\label{sec:max_marginal_likelihood}
MML operates under the assumption that the target text output $y$ is generated by a partially observed random process. Specifically, a text completion $z$ is sampled conditioned on the input $x$ using the language model $\pi_\theta$. Then, each sample $z$ is scored using the true data distribution $P^*(y | x, z)$. This true underlying distribution $P^*$ is generally unknown. The objective of MML is to maximize the log-likelihood of $y$ given $x$, marginalized over all possible $z$: $P_\theta(y | x) = \sum_{z} \pi_\theta (z | x) \times P^*(y | x, z)$.

As the distribution of the generated samples $\pi_\theta(z|x)$ approaches the true distribution $P^*(z|x)$, our objective $P_\theta(y|x)$ will approach the true likelihood $P^*(y | x)$. The key challenge in optimizing MML lies in the computation of the summation over the space of all possible text samples $z$. When training to maximize $\log P_\theta(y | x)$, we are effectively dealing with the log of a sum over all possible text completions, as expressed by the following objective:
\begin{equation}
\label{eq:mml_objective}
\begin{aligned}
J_{\text{MML}}(\theta) = \log \sum_{z} \pi_\theta (z | x) \times P^*(y | x, z)
\end{aligned}
\end{equation}

As explored in earlier work on program synthesis~\cite{guu-etal-2017-language} or and in the recent task of paraphrase generation~\cite{najafi-fyshe-2024-riff}, the MML objective is closely related to the objective of policy gradient methods. Through the application of Jensen's inequality, it can be shown that the MML objective is an upper bound for the policy-gradient objective, yet still MML gradient estimations are more stable~\cite{guu-etal-2017-language, najafi-fyshe-2024-riff}. This relationship can be expressed by the following inequality:
\begin{equation}
\label{eq:mml_pg_ineq}
\begin{aligned}
J_{\text{MML}}(\theta) \ge \sum_{z} \pi_\theta (z | x) \times \log P^*(y | x, z)
\end{aligned}
\end{equation}

\subsection{MML with Automatic Differentiation}
\label{subsec:mml_autodiff}

In practice, optimizing the language model $\pi_{\theta}$ using the MML objective, as presented in Equation~\ref{eq:mml_objective}, requires an approximation due to the intractable summation over all possible samples $z$. To this end, we draw $n$ samples from the conditional distribution $\pi_{\theta}(\cdot|x)$, denoted as $z_1, z_2, \dots, z_n$. The expectation over the sample space is then approximated using numerical summation, which yields the estimate $\hat{J}_{\text{MML}}(\theta)$. To enhance numerical stability during training, it is advantageous to reformulate the objective function in terms of the `log-sum-exp' operation. This reformulation leverages the stable implementation of the `log-sum-exp' function provided by popular machine learning libraries such as PyTorch, TensorFlow, or JAX. In PyTorch, this operation corresponds to the function `torch.logsumexp'\footnote{\url{https://docs.pytorch.org/docs/stable/generated/torch.logsumexp.html}}. This reformulation results in the following approximate objective:

\begin{equation}
\label{eq:mml_pytorch}
\begin{aligned}
\hat{J}_{\text{MML}}(\theta) &= \text{torch.logsumexp}(s_i); \quad 1 \le i \le n \\
s_i &= \log \pi_{\theta}(z_{i} | x) + \log P^{*}(y | x, z_i)
\end{aligned}
\end{equation}

The `log-sum-exp' operation provides numerical stability by mitigating potential issues with exploding exponentials, motivated by the following Lemma:
\begin{lemma}
\label{lemma:logsumexp_stability}
Let $s_1, s_2, \dots, s_n$ be real numbers, and let $s^* = \max_{i} s_i$. Then, $s_i - s^* \le 0$ and we have:
$$ \log \sum_{i=1}^{n} \exp(s_i) = \log \left( \sum_{i=1}^{n} \exp(s_i - s^*)\right) + s^* $$
\end{lemma}

\subsection{MML \& Preference Optimization}
\label{subsec:mml_preference_optimization}

We now articulate our central idea to apply MML estimation for offline preference optimization. Consider a scenario within preference optimization for aligning language models where we have access to only two samples, $z_w$ and $z_l$, for a given input $x$. The RLHF objective can be understood as maximizing two distinct components. The first component is to optimize a new reward function, $R(y, z)$, defined as $R(y, z) = r(y, z) + \beta \log \pi_{\text{sft}}(z|x)$. The second component of the RLHF objective is to maximize the entropy of the language model, $\beta H(\pi_\theta(\cdot|x))$. This connection is summarized by the following equation:
$$J_{\text{RLHF}}(\theta) = \mathbb{E}_{z \sim \pi_\theta(\cdot|x)} \left [ R(y, z) \right ] + \beta H(\pi_\theta(\cdot|x))$$

We first simplify the RLHF objective by ignoring the entropy term. Then, within our MML estimation, we assume that $P^{*}(y | x, z) \propto e^{R(y, z)}$, optimizing the following objective:
\begin{multline}
\label{eq:mml_rlhf_upperbound}
J^{rlhf}_{\text{MML}}(\theta) = \log \sum_{z} \pi_\theta (z | x) \times e^{R(y, z)}, \\
R(y, z) = r(y, z) + \beta \log \pi_{\text{sft}}(z|x)
\end{multline}

The following theorem demonstrates that training the language model using the approximate MML estimation, achieved by optimizing $J^{rlhf}_{\text{MML}}(\theta)$ with the `log-sum-exp' operation is implicitly conducting preference optimization.

\begin{theorem}
\label{theory:implicit_po_mml}
\textbf{MML Estimation on Preference Dataset:} Consider preference data consisting of triplets $(x, z_w, z_l)$, where $x$ is an input, $z_w$ is the preferred response, and $z_l$ is a less preferred response.

When optimizing the approximate MML objective, denoted as $\hat{J}^{rlhf}_{\text{MML}}(\theta)$, and utilizing only the two samples $z_w$ and $z_l$ for the approximation, the resulting gradient update $\nabla_{\theta} \hat{J}^{rlhf}_{\text{MML}}(\theta)$ is equivalent to the sum of two subgradient components: $\nabla^{w}_{\theta} + \nabla^{l}_{\theta}$, defined as:
\begin{equation}
\label{eq:mml_gradient_update}
\begin{aligned}
\nabla^{w}_{\theta} &= \sigma(s_{w} - s_{l}) \nabla_{\theta} \log \pi_{\theta} (z_{w} | x) \\
\nabla^{l}_{\theta} &= \sigma(s_{l} - s_{w}) \nabla_{\theta} \log \pi_{\theta} (z_{l} | x)
\end{aligned}
\end{equation}

In the above subgradient definitions, $\sigma$ represents the sigmoid function. When applying the `logsumexp' operation to approximate the MML objective with the two samples, we inherently assume that the score of the less preferred response, $s_l$, is less than or equal to the score of the preferred response, $s_w$ (i.e., $s_l \le s_w$). These scores are determined by the following equations:
\begin{equation}
\label{eq:mml_gradient_score_def}
\begin{aligned}
s_{w} &=  \log \pi_{\theta}(z_{w} | x) + r(y, z_{w}) + \beta \log \pi_{\text{sft}}(z_{w}|x) \\
s_{l} &=  \log \pi_{\theta}(z_{l} | x) + r(y, z_{l}) + \beta \log \pi_{\text{sft}}(z_{l}|x)
\end{aligned}
\end{equation}
\end{theorem}

The core insight of this theorem (see the appendix~\ref{appendix:proof:implicit_po_mml} for the proof) lies in how the difference in scores between the preferred and less preferred responses modulates the gradient update. The sigmoid function, $\sigma(s_w - s_l)$, plays a crucial role in scaling the gradient component for the preferred response. When the preferred response has a significantly higher score ($s_w \gg s_l$), the sigmoid approaches 1, and the update strongly encourages generating $z_w$. Conversely, the gradient for the less preferred response, $\nabla^{l}_{\theta}$, is scaled by $\sigma(s_l - s_w)$, which is small when $s_l$ is much smaller than $s_w$. This mechanism ensures that the model learns to amplify the probability of the highly preferred response while having a lesser, though non-zero, effect on the less preferred response.

\subsection{MML-based Preference Optimization (MMPO)}
In this section, we describe the loss function for our proposed MMPO objective. We do not have access to the reward function $r(y, z)$, but we aim to enforce the preference constraint $s_{w} \ge s_{l}$. We introduce a hyperparameter $r_\epsilon$ representing the target positive margin for the reward difference, $r(y, z_{w}) - r(y, z_{l})$. Moreover, to provide a better learning signal, we normalize the rewards $R(y,z)$ within a mini-batch of preference pairs. This in-batch normalization stabilizes training, as it ensures the relative scores ($s_{w}, s_{l}$) are on a consistent scale across a mini-batch, preventing outlier examples with unusually high or low reference log-probabilities from dominating the gradient. This kind of reward normalization resembles the normalization of advantage terms in algorithms like PPO~\cite{schulman2017proximalpolicyoptimizationalgorithms}, but is distinct from the per-instance (i.e., per-input) reward normalization used in on-policy algorithms like RLOO~\cite{kool2019buy, ahmadian2024basicsrevisitingreinforcestyle} or GRPO~\cite{shao2024deepseekmathpushinglimitsmathematical}. Since for each input, we have only one preferred ($z_w$) and one dispreferred ($z_l$) completion, we employ in-batch normalization. 

In addition to the $\hat{J}^{rlhf}_{\text{MML}}(\theta)$ objective, we introduce an auxiliary objective to further reinforce the preference condition $s_{w} \ge s_{l}$. While the primary objective implicitly optimizes for preference via its gradient dynamics, we add an auxiliary $\log \sigma(s_{w}(\theta) - s_{l}(\theta))$ term. This component, inspired by DPO's formulation, provides a more direct and explicit supervisory signal to maximize the margin between the scores of the preferred and dis-preferred responses, further stabilizing the learning process. This structural form $\log \sigma(s_w - s_l)$ is similar to the core term in the DPO's loss function, but crucially, it operates on our normalized scores $s_w$ and $s_l$ rather than the log-probability ratios relative to the SFT model used in DPO. 

We have implemented the loss function used for the MMPO objective in our experiments within the PyTorch library. This implementation follows the same coding pattern as the TRL (\url{https://huggingface.co/docs/trl/en/index}) library used for DPO and SimPO. The following code snippet provides this implementation.

\begin{lstlisting}[language=Python, caption=Implementation of MMPO in PyTorch.]
def mmpo_loss(
    self,
    chosen_logps: torch.FloatTensor,
    rejected_logps: torch.FloatTensor,
    ref_chosen_logps: torch.FloatTensor,
    ref_rejected_logps: torch.FloatTensor,
) -> tuple[torch.FloatTensor, torch.FloatTensor, torch.FloatTensor]:
    """
    Compute the MMPO loss for a batch of policy and reference model log probabilities."""

    ep = 1e-6
    b = self.beta  # hyper-param
    r_ep = self.reward_epsilon  # hyper-param

    chosen_rs =  r_ep + b * ref_chosen_logps
    rejected_rs = 0.1 + b * ref_rejected_logps

    # In-batch reward normalization
    all_rs = torch.cat((chosen_rs, rejected_rs), dim=0)
    rs_gather = self.accelerator.gather_for_metrics(all_rs)

    min = rs_gather.min()
    max = rs_gather.max()
    denom = max + ep - min
    norm_chosen_rs = (chosen_rs + ep - min) / denom
    norm_rejected_rs = (rejected_rs + ep - min) / denom

    chosen_s = chosen_logps + norm_chosen_rs
    rejected_s = rejected_logps + norm_rejected_rs

    chosen_s_un = chosen_s.unsqueeze(1)
    rejected_s_un = rejected_s.unsqueeze(1)
    scores = torch.cat((chosen_s_un, rejected_s_un), dim=1)
    losses = -torch.logsumexp(scores, dim=1)
    losses += -F.logsigmoid(chosen_s - rejected_s)

    return losses, chosen_s.detach(), rejected_s.detach()
\end{lstlisting}

\section{Experiments}
\label{experiments}

\begin{figure*}
\begin{center}
\includegraphics[width=1.0\textwidth]{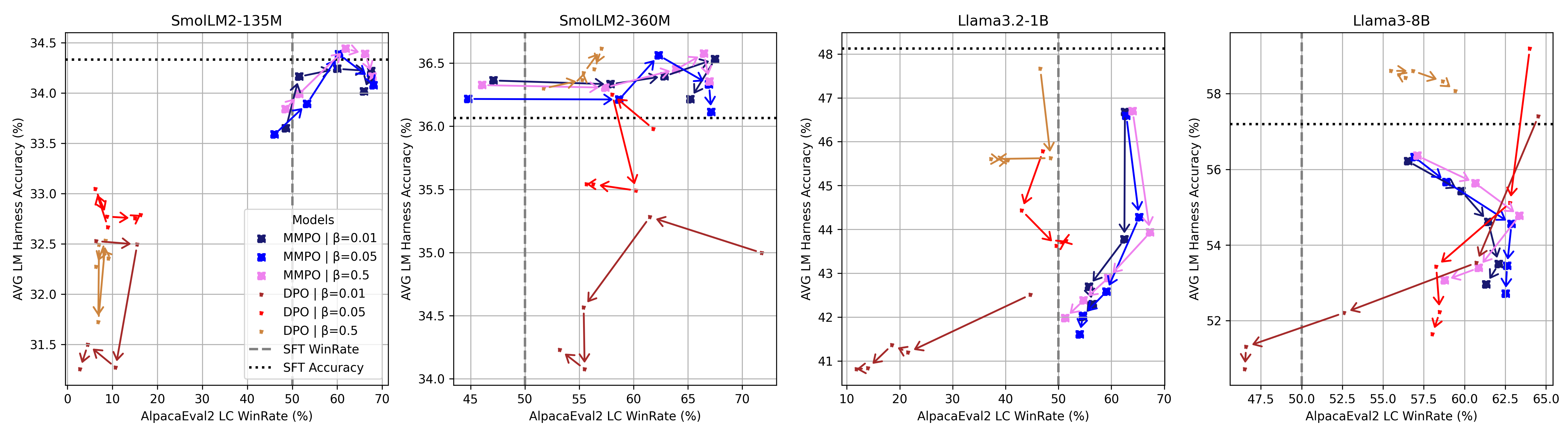}
\end{center}
\caption{The performance trade-off between preference alignment (AlpacaEval2 length-controlled win-rate) and general language capabilities (average LM Harness accuracy). The plots compare MMPO and DPO across four model sizes, with arrows indicating the performance trajectory over five training epochs for various $\beta$ values. Optimal performance would be for the model to retain general language capabilities while increasing preference optimization performance, indicated by lines moving towards the right while falling as little as possible along the y-axis.  Stability across $\beta s$ is achieved when all $\beta$-specific trajectories for a particular model family are clustered together on the graph.}
\label{mmpo-dpo-bubble}
\end{figure*}

\subsection{Setup}

\noindent
\textbf{LM Models:}
\label{lm_models_discuss}
We experimented with two families of language models. First, we instruction-tuned the pre-trained SmolLM2 models (135M and 360M parameters)~\cite{allal2025SmolLM2smolgoesbig} for two epochs on the \texttt{small-smoltalk} dataset\footnote{\url{https://huggingface.co/datasets/HuggingFaceTB/smol-smoltalk}}, which contains approximately 460K examples. Although we adopted the chat template from the publicly available instruction-tuned versions of SmolLM2, we did not use the models themselves because they had already undergone DPO training.

For our second model, Llama3.2-1B, we performed instruction tuning for three epochs on a combination of the \texttt{UltraChat}\footnote{\url{https://huggingface.co/datasets/HuggingFaceH4/ultrachat_200k}} and \texttt{sftdatasetv3}\footnote{\url{https://huggingface.co/datasets/JunxiongWang/sftdatasetv3}} datasets. The latter was recently used to train hybrid Transformer-Mamba models~\cite{junxiongdaniele2024mambainllama}.

Our last model, Llama3-8B, is the publicly available instruction-tuned model trained by the SimPO\footnote{\url{https://huggingface.co/princeton-nlp/Llama-3-Base-8B-SFT}} authors. This model was trained on the \texttt{UltraChat}\footnote{\url{https://huggingface.co/datasets/HuggingFaceH4/ultrachat_200k}} dataset for one epoch~\cite{meng2024simpo}.\footnote{These LM models were pre-trained for a varying number of epochs, depending on resource availability.}

\noindent
\textbf{Preference Datasets:}
\label{dataset_discuss}
For the preference optimization of our instruction-tuned models, SmolLM2 (135M and 360M), Llama3.2-1B and Llama3-8B, we combined two publicly available datasets. The first was the \texttt{Orca DPO Pairs} dataset~\cite{OpenOrca}, which contains 12,359 training and 500 test preference pairs\footnote{\url{https://huggingface.co/datasets/HuggingFaceH4/orca_dpo_pairs}}. The second was the \texttt{UltraFeedback-Binarized} dataset~\cite{cui2023ultrafeedback}, containing 61,135 training and 2,000 test preference pairs\footnote{\url{https://huggingface.co/datasets/HuggingFaceH4/ultrafeedback_binarized}}. In the latter, preference pairs were generated by using GPT-4 to score and select completions for prompts from the UltraFeedback dataset.

\noindent
\textbf{Baselines:}
\label{baseline_description}
We consider two baselines in our experiments. We compare against DPO, which optimizes a logistic objective of the log-probability ratios (see Appendix~\ref{appendix:methods_dpo} for the exact objective). Our second baseline is the recent SimPO method~\cite{meng2024simpo}, which adopts a similar pairwise logistic objective but removes the reference policy, using only the model’s own log-probabilities:
\begin{equation}
\label{eq:simpo_objective}
\begin{aligned}
\log \sigma\left(\beta \log \pi_\theta(z_w|x) - \beta \log \pi_\theta(z_l|x) - \gamma\right)
\end{aligned}
\end{equation}
By eliminating the reference model, SimPO is computationally lighter than DPO but relies on $\beta$ and the margin $\gamma$ to regulate distributional drift compared to the initial reference model.

\noindent
\textbf{Hyperparameters:}
\label{hyper_param_discuss}
We tuned the hyperparameters for the preference optimization stage using a multi-stage grid search for each model size (135M, 360M, 1B, and 8B). To select the best-performing hyperparameters, we used the reward accuracy metric on the validation split of our preference datasets. This metric evaluates how frequently the internal score assigned to a chosen response is higher than the score for the corresponding rejected response. We observed that the reward accuracy metric consistently favored the smallest value, $\beta=0.01$, across all models. To provide a comprehensive analysis of the objectives under different $\beta$ settings, we subsequently trained the models for five epochs for each value of $\beta \in \{0.01, 0.05, 0.5\}$.
Further optimization and training details are outlined in Appendix~\ref{appendix:experiments:hyper}.

\subsection{Results}
\subsubsection{Language Evaluation}

\begin{figure*}
\begin{center}
\includegraphics[width=1.0\textwidth]{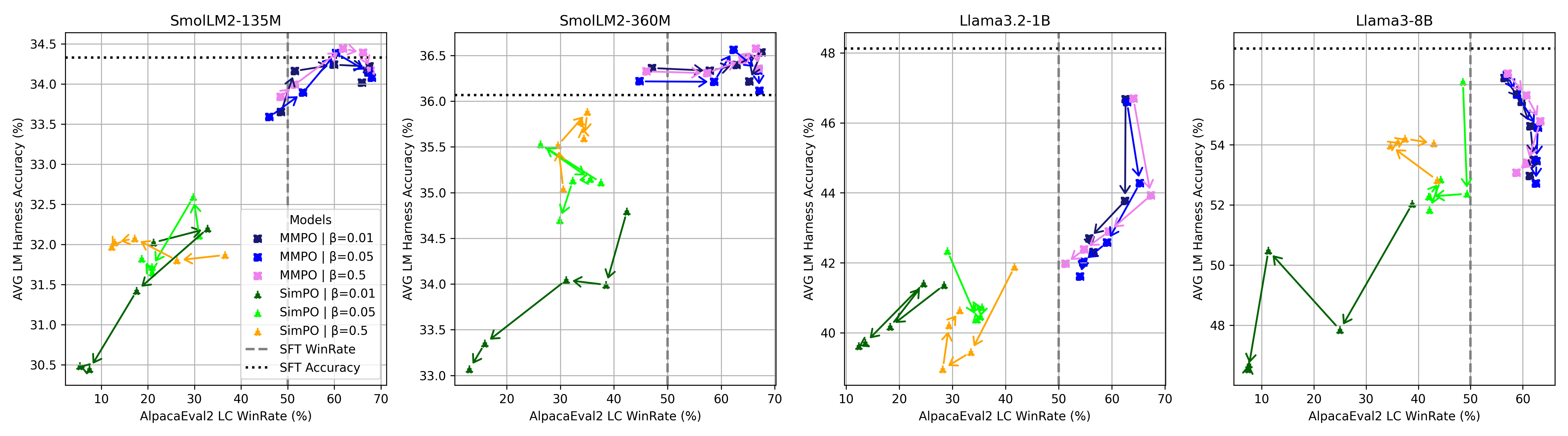}
\end{center}
\caption{The performance trade-off between preference alignment (AlpacaEval2 length-controlled win-rate) and general language capabilities (average LM Harness accuracy). The plots compare MMPO and SimPO across four model sizes, with arrows indicating the performance trajectory over five training epochs for various $\beta$ values.}
\label{mmpo-simpo-bubble}
\end{figure*}

One of our key questions is the extent to which the preference optimization impacts the general capabilities acquired during the pre-training and instruction-tuning stages, particularly given that these earlier phases use orders of magnitude more data. While a specific preference optimization method may improve a model's ability to distinguish between preferred and dispreferred responses, it is crucial to ensure this does not cause a degradation in the model's broader language understanding and reasoning abilities.

To assess the general capabilities of our models, we employ the Language Model Evaluation Harness benchmark~\citep{gao2021framework}. We evaluate the models in a zero-shot setting on the test splits of 10 diverse language understanding tasks: \texttt{winogrande}, \texttt{arc\_easy}, \texttt{piqa}, \texttt{hellaswag}, \texttt{openbookqa}, \texttt{arc\_challenge}, \texttt{mmlu}, \texttt{mathqa}, \texttt{race}, and \texttt{commonsense\_qa}~\citep{sakaguchi2019winograndeadversarialwinogradschema, clark2018thinksolvedquestionanswering, bisk2019piqareasoningphysicalcommonsense, zellers2019hellaswagmachinereallyfinish, mihaylov-etal-2018-suit, hendrycks2021measuringmassivemultitasklanguage, amini-etal-2019-mathqa, lai-etal-2017-race, talmor-etal-2019-commonsenseqa}. These datasets cover multiple question-answering and commonsense reasoning tasks. Our final reported metric is the average accuracy across these ten tasks.

Our experiments, summarized in y-axis of Figures~\ref{mmpo-dpo-bubble} and~\ref{mmpo-simpo-bubble}, compare MMPO with DPO and SimPO on the LM Harness benchmark. A key finding is the stability of MMPO with respect to the hyperparameter $\beta$. Unlike MMPO, both DPO and SimPO show significant sensitivity to $\beta$. We hypothesize that MMPO's stability is primarily due to in-batch normalization of the scores and implicit subtraction of the scores within the gradients.

Furthermore, MMPO proved to be more effective at preserving the general language understanding and reasoning capabilities of instruction-tuned models. While all preference optimization methods can cause some performance degradation at the end of epoch five, this impact is minimal with MMPO (corresponding to $\beta=0.01$). In Llama-3-8B, the tuned learning rate for DPO is 10 times smaller than the tuned learning rate for MMPO, however, after three epochs, we still see more performance degradation compared to the MMPO (corresponding to $\beta=0.01$).

\subsubsection{Preference Evaluation}

For preference evaluation, we use the AlpacaEval-2 benchmark, which consists of 805 diverse questions~\cite{alpaca_eval}. For our evaluation, we compute the win-rate of our models against the baseline instruction-tuned model. The comparison is performed by two LLM-as-judges, which evaluate pairs of responses to determine the preferred one. Our first judge is ``weighted\_alpaca\_eval\_gpt-4o-mini-2024-07-18'', and the second judge is ``claude-haiku-4-5-20251001''. To mitigate the known tendency of LLM judges to favor longer outputs regardless of quality, we report the length-controlled win-rate, which normalizes for this bias. This metric reflects the percentage of times our model's response is judged superior to the baseline's, focusing purely on instruction adherence and helpfulness~\cite{dubois2024length}\footnote{We used the github code for AlpacaEval 2.0 benchmark. More information about the templates used within judges are provided by the benchmark itself.}. Results from these two judges are averaged and summarized in x-axis of Figures~\ref{mmpo-dpo-bubble} and ~\ref{mmpo-simpo-bubble}.

MMPO demonstrates stability under hyperparameter tuning, with negligible performance variation across a range of $\beta$ values. For instance, with the SmolLM2-135M model, MMPO's win-rate improves monotonically from 47.7\% at epoch 1 to a final rate of 67.3\% by epoch 5 (averaged across all $\beta$ values). This stable behavior is in stark contrast to DPO, which yields very low win-rates, and SimPO, which degrades from an early peak of about 37\% to 5–20\%.

\begin{figure*}
\begin{center}
\includegraphics[width=1.0\textwidth,]{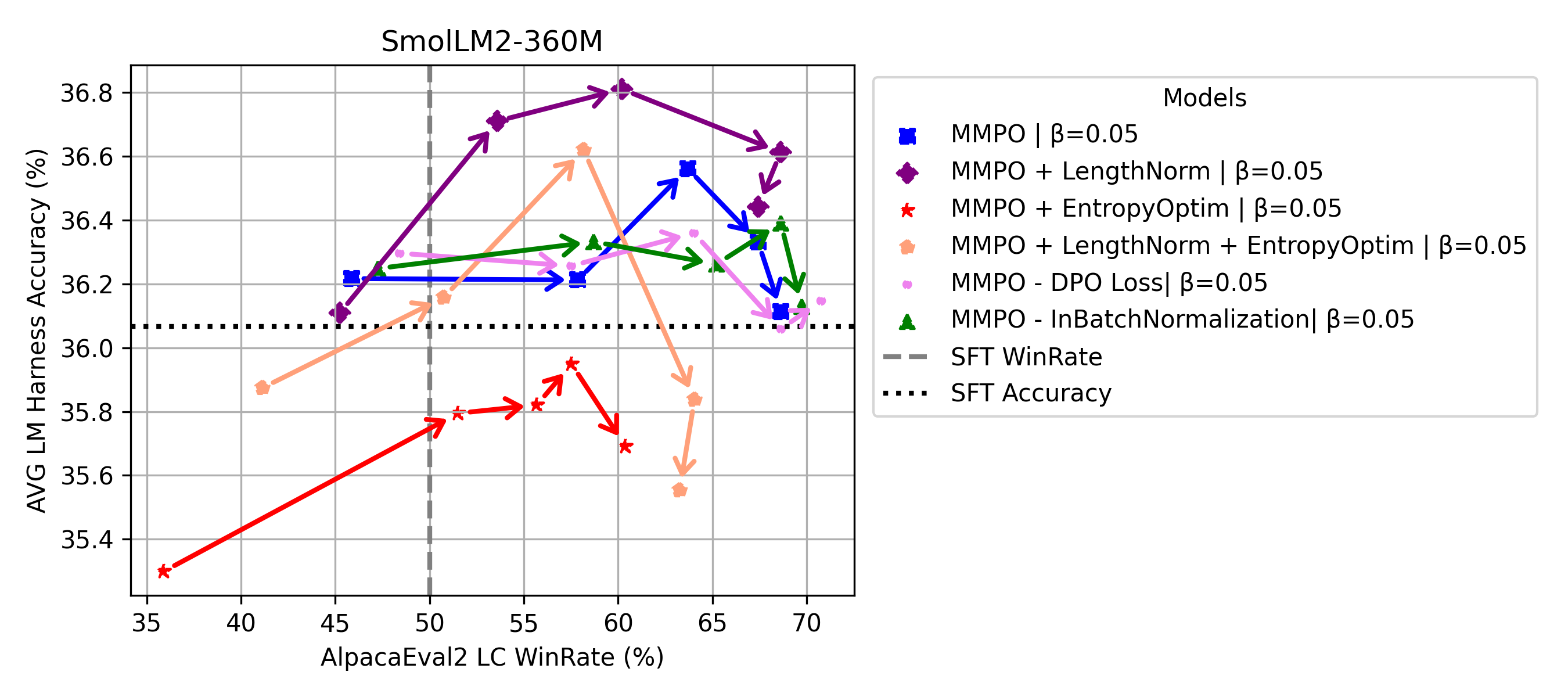}
\end{center}
\caption{An ablation study of the MMPO objective ($\beta=0.05$) on the SmolLM2-360M model. The plot evaluates the trade-off between preference alignment, measured by the AlpacaEval2 length-controlled (LC) win-rate, and general language capabilities, measured by the average LM Harness accuracy. The tested modifications include removing the DPO-style auxiliary loss, removing in-batch normalization, adding length normalization, and adding entropy maximization. Arrows indicate the performance trajectory over five training epochs.}
\label{mmpo-ablation-results}
\end{figure*}

With the SmolLM2-360M, MMPO again climbs steadily to 65–67\% while remaining highly stable to $\beta$. In contrast, DPO shows strong $\beta$-sensitivity, starting at a high of 71.8\% for $\beta=0.01$ but then drifting downward to 53.2\% by epoch 5. SimPO's performance tops out near 41\% and is otherwise lower. While DPO with $\beta=0.01$ achieves the highest single win-rate on the AlpacaEval-2 benchmark, this performance comes at a significant cost, corresponding to a 1.5\% average performance drop on the LM Harness benchmark compared to the MMPO (see DPO's performance at epoch 1 in Figure~\ref{mmpo-dpo-bubble}).

On the Llama3.2-1B model, MMPO reaches 62–67\% within two epochs and settles at 51–54\% by epoch 5. DPO, however, proves highly unstable; for example, with $\beta=0.01$, DPO collapses from around 44.7\% to 11.8\%, with only the setting $\beta=0.05$ approaching MMPO's performance late in training (51.4\%). SimPO consistently remains far below MMPO across all $\beta$ values. On the Llama3.2-1B model, we observe that with only 1 epoch of MMPO training with any of the beta values, we achieve a 62\% win-rate with minimally sacrificing the LM harness accuracy.

In Llama3-8B, MMPO with $\beta=0.5$ achieves the win-rate of 63.4\% by the end of epoch 4, which is competitive to DPO with the highest win rate of 64.5\% ($\beta=0.01$).

Across the models SmolLM2-135M, SmolLM2-360M, and Llama3.2-1B, MMPO offers a more favorable Pareto profile: it delivers consistently higher preference win-rates with minimal degradation in general capabilities.
Across all model sizes, MMPO is largely invariant to $\beta$. DPO can achieve strong early win-rates under carefully tuned $\beta$, but at the cost of instability with respect to $\beta$. SimPO is generally less competitive and less stable according to our experiments\footnote{The original SimPO paper tested the objective only with models having 8B parameters under full-parameter training; therefore, it's unclear if SimPO's superior performance over DPO holds true in smaller models with limited capacity or under parameter-efficient training.}.

Tables~\ref{peak-lm-harness} and \ref{peak-winrate} summarize the relationship between LM Harness accuracy and the averaged length-controlled win-rates (computed across two judges). Table~\ref{peak-lm-harness} reports the win-rate observed at the point of peak LM Harness accuracy (across all $\beta$ values), while Table~\ref{peak-winrate} reports the LM Harness accuracy observed at the point of peak win-rate (across all $\beta$ values). Please note that win-rates are computed against the SFT baseline of the respective model size and are not comparable across different model sizes.

For smaller model sizes (135M, 360M, and 1B), MMPO demonstrates a superior Pareto profile. We note that our 8B experiments were conducted under a more constrained parameter-efficient setting, using the 8-bit Adam optimizer and applying LoRA exclusively to attention projection matrices. Investigating full-parameter training with the 8B model remains a promising direction for future work.

\begin{table}[t]
    \centering
    \small
    \renewcommand{\arraystretch}{1.1}
    \caption{Comparison of peak LM Harness accuracies and corresponding AlpacaEval-2 win-rates across model sizes.}
    \label{tab:model_performance}
    \begin{tabular}{l l c c}
    \toprule
    \textbf{Model Size} & \textbf{Method} & \textbf{Peak LM Acc} & \textbf{Win-Rate} \\
    \midrule
    \multirow{2}{*}{SmolLM 135M} & MMPO & \textbf{34.44} & \textbf{61.86} \\
    & DPO & 33.05 & 6.21 \\ 
    \midrule
    \multirow{2}{*}{SmolLM 360M} & MMPO & 36.58 & \textbf{66.47} \\
    & DPO  & \textbf{36.62} & 57.06 \\ 
    \midrule
    \multirow{2}{*}{Llama3 1B} & MMPO & 46.69 & \textbf{64.03} \\
    & DPO  & \textbf{47.67} & 46.57 \\ 
    \midrule
    \multirow{2}{*}{Llama3 8B} & MMPO & 56.36 & 57.11 \\
     & DPO  & \textbf{59.19} & \textbf{64.01} \\
    \bottomrule
    \end{tabular}
    \label{peak-lm-harness}
\end{table}

\begin{table}[t]
    \centering
    \small
    \renewcommand{\arraystretch}{1.1}
    \caption{Comparison of peak AlpacaEval-2 win-rates and corresponding LM Harness accuracies across model sizes.}
    \label{tab:peak_win_rate}
    \begin{tabular}{l l c c}
    \toprule
    \textbf{Model Size} & \textbf{Method} & \textbf{Peak Win-Rate} & \textbf{LM Acc} \\
    \midrule
    \multirow{2}{*}{SmolLM 135M} & MMPO & \textbf{68.09} & \textbf{34.08} \\
        & DPO & 16.29 & 32.79 \\ 
    \midrule
    \multirow{2}{*}{SmolLM 360M} & MMPO & 67.50 & \textbf{36.53} \\
        & DPO & \textbf{71.80} & 35.00 \\ 
    \midrule
    \multirow{2}{*}{Llama3 1B} & MMPO & \textbf{67.32} & \textbf{43.93} \\
        & DPO & 51.45 & 43.74 \\ 
    \midrule
    \multirow{2}{*}{Llama3 8B} & MMPO & 63.37 & 54.77 \\
        & DPO  & \textbf{64.53} & \textbf{57.40} \\
    \bottomrule
    \end{tabular}
    \label{peak-winrate}
\end{table}

\section{Ablation Experiments}
We conducted four ablation studies on the SmolLM2-360M model to systematically investigate the contribution of key components within the MMPO objective. The experiments focused on: (1) \textbf{Length Normalization}, (2) \textbf{Entropy Maximization}, (3) the \textbf{DPO-style Auxiliary Objective}, and (4) \textbf{In-batch Normalization}. All studies were performed with a $\beta$ value of $0.05$, a setting that demonstrated robust performance on the LM-Harness and AlpacaEval-2 benchmarks.

\noindent
\textbf{Length Normalization:}
For our first ablation study, we modified the MMPO objective to normalize the total log-probability of a sequence by the number of tokens, thereby optimizing for the average log-probability per token.

\vspace{0.5em}
\noindent
\textbf{Entropy Maximization:}
We studied the impact of entropy maximization. We reintroduced the entropy term, $\beta H(\pi_\theta(\cdot|x))$, which was excluded from our final objective, to explicitly encourage a more diverse policy. The gradient of this term was approximated using a token-level entropy objective at each generation step, as detailed in Appendix~\ref{appendix:experiments:ablation_entropy}.

\vspace{0.5em}
\noindent
\textbf{DPO-style Auxiliary Objective:}
Our third study evaluated the necessity of the DPO-style auxiliary objective, $\log \sigma(s_{w}(\theta) - s_{l}(\theta))$. By removing this term, we determine if the main `logsumexp' term in the MMPO loss is sufficient on its own for effective preference optimization.

\vspace{0.5em}
\noindent
\textbf{In-batch Normalization:}
Finally, we explored the importance of in-batch normalization. This ablation involved removing the normalization step from the MMPO objective and relying solely on the unnormalized scores $s_w$ and $s_l$. The goal was to understand if this normalization is a critical component for the method's success or if the implicit preference optimization from the main objective is the dominant factor driving performance gains.

Figure~\ref{mmpo-ablation-results} summarizes the results for our ablation experiments:

\begin{itemize}
    \item \textbf{Length Normalization} improves the average LM Harness accuracy, but it does not lead to a significant improvement in AlpacaEval-2 win-rates. This suggests that while it may help the model to learn more accurate token probabilities, it does not translate directly to more helpful or preferred responses.
    \item In contrast, \textbf{Entropy Maximization} consistently degrades performance on both the LM Harness and AlpacaEval-2 benchmarks, indicating that explicitly encouraging language model diversity via this method is detrimental to our objective. Note that explicit entropy regularization, approximated as token-level entropy, is not helping our objective. This observation does not prove that the full KL-regularization used in RLHF is irrelevant. The scores from the reference model still contribute to the scores $s_w$ and $s_l$ within the MML objective or even the DPO-style auxiliary loss.
 
    \item Furthermore, removing the \textbf{DPO-style Auxiliary Objective} or \textbf{In-batch Normalization} leads to a drop in peak LM Harness accuracy but a slight improvement in AlpacaEval-2 win rates. This intriguing result suggests that the implicit preference learning signal from the core \texttt{logsumexp} term is a strong driver of performance on its own. These ablation results illustrate a trade-off rather than a fundamental flaw in the objective; As discussed, while omitting the DPO-style auxiliary loss (``MMPO - DPO Loss'') or in-batch normalization slightly improves AlpacaEval-2 win rates, it simultaneously degrades peak LM Harness accuracy.
\end{itemize}

\section{Related Works}
Numerous methods have generalized DPO~\cite{rafailov2023direct}. The method f-DPO~\cite{wang2023fdpo} broadens the objective by replacing the reverse-KL regularizer in RLHF with a more general class of f-divergences, allowing for different constraints on model deviation. Generalized Preference Optimization (GPO)~\cite{tang2024gpo} provides a unifying perspective, showing that methods like DPO can be seen as specific instances of optimizing convex contrastive losses. More recently, Calibrated DPO (Cal-DPO)~\cite{xiao2024caldpo} addresses DPO's training instability by explicitly calibrating the implicit reward scale ($\beta$ parameter) to achieve more robust performance.

MMPO is fundamentally different from this family of methods, as it is not a contrastive surrogate for a reward maximization problem. The gradient of MMPO produces an implicit, data-dependent weighting of the chosen and rejected responses, rather than relying on an explicit, fixed divergence constraint. Furthermore, MMPO achieves stable training without the explicit mechanisms of Cal-DPO. Its stability is an inherent property of the formulation, stemming from in-batch score normalization and the natural subtraction within the `logsumexp' operation, which prevents scores from growing unbounded.

Another category of methods avoids online preference optimization by leveraging ranking or data filtering. Rank Responses to Align Human Feedback (RRHF)~\cite{yuan2023rrhf} trains a model using a ranking loss over multiple candidate responses for a given prompt, encouraging it to assign higher likelihoods to better responses. Reward-Ranked Fine-Tuning (RAFT)~\cite{dong2023raft} uses an external reward model to score and rank multiple model-generated responses, then performs standard supervised fine-tuning on a filtered dataset containing only the top-ranked response for each prompt. MMPO’s approach is distinct from both. Unlike RRHF, it operates on pairwise preference data without requiring multi-candidate rankings. Unlike RAFT, it is self-contained and does not depend on an external reward model to curate or filter training data.

For our experiments, we chose to compare MMPO against well-tuned DPO and SimPO. This decision was based on the findings by the SimPO authors that a well-tuned DPO can outperform many alternative preference optimization techniques such as RRHF. It is worth noting, however, that we were unable to replicate the reported superior performance of SimPO over DPO. In all of our experiments, DPO consistently outperforms SimPO.
\section{Conclusion}
We introduced the MMPO objective, a novel and simple objective for offline LLM alignment. Our approach treats preference pairs as samples to approximate the marginal likelihood of generating the desirable output given the input text, which we have shown theoretically to be equivalent to an implicit preference optimization scheme that requires neither an explicit reward function nor an entropy bonus.

Our extensive experiments on models up to 8B parameters demonstrate MMPO's key advantages. It exhibits stability with respect to the hyperparameter $\beta$, a common source of instability in methods like DPO and SimPO. Crucially, MMPO achieves competitive preference alignment score, as measured by win-rates on AlpacaEval-2 benchmark, while causing less degradation to the model's general language understanding and reasoning abilities on the LM Harness benchmark.

\section{Limitations}
A limitation of our current experimental setup is the inability to directly evaluate our preference optimization methods on models exceeding 1 billion parameters without employing memory-efficiency techniques such as LoRA or 8-bit optimizers. Testing these approaches on larger-scale models is crucial for understanding their full potential and scalability. This was primarily constrained by access to computing resources with sufficient GPU memory. Investigating the performance of MMPO on larger models, potentially updating all the parameters using full-precision optimizers, remains an interesting direction for future work.

Our proposed MMPO algorithm is presented and evaluated in the context of offline preference optimization using a static dataset. While MMPO is designed for this offline setting, it could potentially be adapted for online preference optimization. This would likely involve incorporating access to dynamic reward functions and potentially sampling multiple responses per input during training. Exploring the application of MMPO to online scenarios and conducting comparisons with established online methods like PPO or GRPO represents another promising avenue for future research.

\bibliography{custom.bib}

\begin{thebibliography}{40}
\providecommand{\natexlab}[1]{#1}

\bibitem[{Ahmadian et~al.(2024)Ahmadian, Cremer, Gallé, Fadaee, Kreutzer,
  Pietquin, Üstün, and Hooker}]{ahmadian2024basicsrevisitingreinforcestyle}
Arash Ahmadian, Chris Cremer, Matthias Gallé, Marzieh Fadaee, Julia Kreutzer,
  Olivier Pietquin, Ahmet Üstün, and Sara Hooker. 2024.
\newblock \href {https://arxiv.org/abs/2402.14740} {Back to basics: Revisiting
  reinforce style optimization for learning from human feedback in llms}.
\newblock \emph{Preprint}, arXiv:2402.14740.

\bibitem[{Allal et~al.(2025)Allal, Lozhkov, Bakouch, Blázquez, Penedo,
  Tunstall, Marafioti, Kydlíček, Lajarín, Srivastav, Lochner, Fahlgren,
  Nguyen, Fourrier, Burtenshaw, Larcher, Zhao, Zakka, Morlon, Raffel, von
  Werra, and Wolf}]{allal2025SmolLM2smolgoesbig}
Loubna~Ben Allal, Anton Lozhkov, Elie Bakouch, Gabriel~Martín Blázquez,
  Guilherme Penedo, Lewis Tunstall, Andrés Marafioti, Hynek Kydlíček,
  Agustín~Piqueres Lajarín, Vaibhav Srivastav, Joshua Lochner, Caleb
  Fahlgren, Xuan-Son Nguyen, Clémentine Fourrier, Ben Burtenshaw, Hugo
  Larcher, Haojun Zhao, Cyril Zakka, Mathieu Morlon, Colin Raffel, Leandro von
  Werra, and Thomas Wolf. 2025.
\newblock \href {https://arxiv.org/abs/2502.02737} {Smollm2: When smol goes big
  -- data-centric training of a small language model}.
\newblock \emph{Preprint}, arXiv:2502.02737.

\bibitem[{Amini et~al.(2019)Amini, Gabriel, Lin, Koncel-Kedziorski, Choi, and
  Hajishirzi}]{amini-etal-2019-mathqa}
Aida Amini, Saadia Gabriel, Shanchuan Lin, Rik Koncel-Kedziorski, Yejin Choi,
  and Hannaneh Hajishirzi. 2019.
\newblock \href {https://doi.org/10.18653/v1/N19-1245} {{M}ath{QA}: Towards
  interpretable math word problem solving with operation-based formalisms}.
\newblock In \emph{Proceedings of the 2019 Conference of the North {A}merican
  Chapter of the Association for Computational Linguistics: Human Language
  Technologies, Volume 1 (Long and Short Papers)}, pages 2357--2367,
  Minneapolis, Minnesota. Association for Computational Linguistics.

\bibitem[{Bisk et~al.(2019)Bisk, Zellers, Bras, Gao, and
  Choi}]{bisk2019piqareasoningphysicalcommonsense}
Yonatan Bisk, Rowan Zellers, Ronan~Le Bras, Jianfeng Gao, and Yejin Choi. 2019.
\newblock \href {https://arxiv.org/abs/1911.11641} {Piqa: Reasoning about
  physical commonsense in natural language}.
\newblock \emph{Preprint}, arXiv:1911.11641.

\bibitem[{Clark et~al.(2018)Clark, Cowhey, Etzioni, Khot, Sabharwal, Schoenick,
  and Tafjord}]{clark2018thinksolvedquestionanswering}
Peter Clark, Isaac Cowhey, Oren Etzioni, Tushar Khot, Ashish Sabharwal, Carissa
  Schoenick, and Oyvind Tafjord. 2018.
\newblock \href {https://arxiv.org/abs/1803.05457} {Think you have solved
  question answering? try arc, the ai2 reasoning challenge}.
\newblock \emph{Preprint}, arXiv:1803.05457.

\bibitem[{Cui et~al.(2023)Cui, Yuan, Ding, Yao, Zhu, Ni, Xie, Liu, and
  Sun}]{cui2023ultrafeedback}
Ganqu Cui, Lifan Yuan, Ning Ding, Guanming Yao, Wei Zhu, Yuan Ni, Guotong Xie,
  Zhiyuan Liu, and Maosong Sun. 2023.
\newblock \href {https://arxiv.org/abs/2310.01377} {Ultrafeedback: Boosting
  language models with high-quality feedback}.
\newblock \emph{Preprint}, arXiv:2310.01377.

\bibitem[{Dettmers et~al.(2022)Dettmers, Lewis, Shleifer, and
  Zettlemoyer}]{dettmers20228bitoptimizersblockwisequantization}
Tim Dettmers, Mike Lewis, Sam Shleifer, and Luke Zettlemoyer. 2022.
\newblock \href {https://arxiv.org/abs/2110.02861} {8-bit optimizers via
  block-wise quantization}.
\newblock \emph{Preprint}, arXiv:2110.02861.

\bibitem[{Dong et~al.(2023)Dong, Xiong, Goyal, Zhang, Chow, Pan, Diao, Zhang,
  Shum, and Zhang}]{dong2023raft}
Hanze Dong, Wei Xiong, Deepanshu Goyal, Yihan Zhang, Winnie Chow, Rui Pan,
  Shizhe Diao, Jipeng Zhang, Kashun Shum, and Tong Zhang. 2023.
\newblock {RAFT}: Reward ranked finetuning for generative foundation model
  alignment.
\newblock \emph{arXiv preprint arXiv:2304.06767}.

\bibitem[{Dubois et~al.(2024)Dubois, Galambosi, Liang, and
  Hashimoto}]{dubois2024length}
Yann Dubois, Bal{\'a}zs Galambosi, Percy Liang, and Tatsunori~B Hashimoto.
  2024.
\newblock Length-controlled alpacaeval: A simple way to debias automatic
  evaluators.
\newblock \emph{arXiv preprint arXiv:2404.04475}.

\bibitem[{Engstrom et~al.(2020)Engstrom, Ilyas, Santurkar, Tsipras, Janoos,
  Rudolph, and Madry}]{engstrom2020implementationmattersdeeppolicy}
Logan Engstrom, Andrew Ilyas, Shibani Santurkar, Dimitris Tsipras, Firdaus
  Janoos, Larry Rudolph, and Aleksander Madry. 2020.
\newblock \href {https://arxiv.org/abs/2005.12729} {Implementation matters in
  deep policy gradients: A case study on ppo and trpo}.
\newblock \emph{Preprint}, arXiv:2005.12729.

\bibitem[{Gao et~al.(2021)Gao, Tow, Abbasi, Biderman, Black, DiPofi, Foster,
  Gold, Hsu, Leahy et~al.}]{gao2021framework}
Leo Gao, Jonathan Tow, Bena Abbasi, Stella Biderman, Sid Black, Anthony DiPofi,
  Charles Foster, Laurence Gold, Jeffrey Hsu, Kyle Leahy, et~al. 2021.
\newblock A framework for evaluating language models.
\newblock In \emph{2021 NeurIPS Workshop on Datasets and Benchmarks}.

\bibitem[{Guu et~al.(2017)Guu, Pasupat, Liu, and
  Liang}]{guu-etal-2017-language}
Kelvin Guu, Panupong Pasupat, Evan Liu, and Percy Liang. 2017.
\newblock \href {https://doi.org/10.18653/v1/P17-1097} {From language to
  programs: Bridging reinforcement learning and maximum marginal likelihood}.
\newblock In \emph{Proceedings of the 55th Annual Meeting of the Association
  for Computational Linguistics (Volume 1: Long Papers)}, pages 1051--1062,
  Vancouver, Canada. Association for Computational Linguistics.

\bibitem[{Hendrycks et~al.(2021)Hendrycks, Burns, Basart, Zou, Mazeika, Song,
  and Steinhardt}]{hendrycks2021measuringmassivemultitasklanguage}
Dan Hendrycks, Collin Burns, Steven Basart, Andy Zou, Mantas Mazeika, Dawn
  Song, and Jacob Steinhardt. 2021.
\newblock \href {https://arxiv.org/abs/2009.03300} {Measuring massive multitask
  language understanding}.
\newblock \emph{Preprint}, arXiv:2009.03300.

\bibitem[{Hu et~al.(2021)Hu, Shen, Wallis, Allen-Zhu, Li, Wang, Wang, and
  Chen}]{hu2021loralowrankadaptationlarge}
Edward~J. Hu, Yelong Shen, Phillip Wallis, Zeyuan Allen-Zhu, Yuanzhi Li, Shean
  Wang, Lu~Wang, and Weizhu Chen. 2021.
\newblock \href {https://arxiv.org/abs/2106.09685} {Lora: Low-rank adaptation
  of large language models}.
\newblock \emph{Preprint}, arXiv:2106.09685.

\bibitem[{Ilyas et~al.(2020)Ilyas, Engstrom, Santurkar, Tsipras, Janoos,
  Rudolph, and Madry}]{ilyas2020closerlookdeeppolicy}
Andrew Ilyas, Logan Engstrom, Shibani Santurkar, Dimitris Tsipras, Firdaus
  Janoos, Larry Rudolph, and Aleksander Madry. 2020.
\newblock \href {https://arxiv.org/abs/1811.02553} {A closer look at deep
  policy gradients}.
\newblock \emph{Preprint}, arXiv:1811.02553.

\bibitem[{Kool et~al.(2019)Kool, van Hoof, and Welling}]{kool2019buy}
Wouter Kool, Herke van Hoof, and Max Welling. 2019.
\newblock \href {https://openreview.net/forum?id=r1lgTGL5DE} {Buy 4 {REINFORCE}
  samples, get a baseline for free!}

\bibitem[{Lai et~al.(2017)Lai, Xie, Liu, Yang, and Hovy}]{lai-etal-2017-race}
Guokun Lai, Qizhe Xie, Hanxiao Liu, Yiming Yang, and Eduard Hovy. 2017.
\newblock \href {https://doi.org/10.18653/v1/D17-1082} {{RACE}: Large-scale
  {R}e{A}ding comprehension dataset from examinations}.
\newblock In \emph{Proceedings of the 2017 Conference on Empirical Methods in
  Natural Language Processing}, pages 785--794, Copenhagen, Denmark.
  Association for Computational Linguistics.

\bibitem[{Li et~al.(2023)Li, Zhang, Dubois, Taori, Gulrajani, Guestrin, Liang,
  and Hashimoto}]{alpaca_eval}
Xuechen Li, Tianyi Zhang, Yann Dubois, Rohan Taori, Ishaan Gulrajani, Carlos
  Guestrin, Percy Liang, and Tatsunori~B. Hashimoto. 2023.
\newblock Alpacaeval: An automatic evaluator of instruction-following models.
\newblock \url{https://github.com/tatsu-lab/alpaca_eval}.

\bibitem[{Lian et~al.(2023)Lian, Goodson, Pentland, Cook, Vong, and
  "Teknium"}]{OpenOrca}
Wing Lian, Bleys Goodson, Eugene Pentland, Austin Cook, Chanvichet Vong, and
  "Teknium". 2023.
\newblock Openorca: An open dataset of gpt augmented flan reasoning traces.
\newblock \url{https://https://huggingface.co/Open-Orca/OpenOrca}.

\bibitem[{Loshchilov and
  Hutter(2017)}]{loshchilov2017sgdrstochasticgradientdescent}
Ilya Loshchilov and Frank Hutter. 2017.
\newblock \href {https://arxiv.org/abs/1608.03983} {Sgdr: Stochastic gradient
  descent with warm restarts}.
\newblock \emph{Preprint}, arXiv:1608.03983.

\bibitem[{Loshchilov and
  Hutter(2019)}]{loshchilov2019decoupledweightdecayregularization}
Ilya Loshchilov and Frank Hutter. 2019.
\newblock \href {https://arxiv.org/abs/1711.05101} {Decoupled weight decay
  regularization}.
\newblock \emph{Preprint}, arXiv:1711.05101.

\bibitem[{Meng et~al.(2024)Meng, Xia, and Chen}]{meng2024simpo}
Yu~Meng, Mengzhou Xia, and Danqi Chen. 2024.
\newblock Simpo: Simple preference optimization with a reference-free reward.
\newblock In \emph{Advances in Neural Information Processing Systems
  (NeurIPS)}.

\bibitem[{Mihaylov et~al.(2018)Mihaylov, Clark, Khot, and
  Sabharwal}]{mihaylov-etal-2018-suit}
Todor Mihaylov, Peter Clark, Tushar Khot, and Ashish Sabharwal. 2018.
\newblock \href {https://doi.org/10.18653/v1/D18-1260} {Can a suit of armor
  conduct electricity? a new dataset for open book question answering}.
\newblock In \emph{Proceedings of the 2018 Conference on Empirical Methods in
  Natural Language Processing}, pages 2381--2391, Brussels, Belgium.
  Association for Computational Linguistics.

\bibitem[{Najafi and Fyshe(2024)}]{najafi-fyshe-2024-riff}
Saeed Najafi and Alona Fyshe. 2024.
\newblock \href {https://doi.org/10.18653/v1/2024.findings-acl.85} {{RIFF}:
  Learning to rephrase inputs for few-shot fine-tuning of language models}.
\newblock In \emph{Findings of the Association for Computational Linguistics:
  ACL 2024}, pages 1447--1466, Bangkok, Thailand. Association for Computational
  Linguistics.

\bibitem[{Ouyang et~al.(2022)Ouyang, Wu, Jiang, Almeida, Wainwright, Mishkin,
  Zhang, Agarwal, Slama, Gray, Schulman, Hilton, Kelton, Miller, Simens,
  Askell, Welinder, Christiano, Leike, and Lowe}]{ouyang2022training}
Long Ouyang, Jeffrey Wu, Xu~Jiang, Diogo Almeida, Carroll Wainwright, Pamela
  Mishkin, Chong Zhang, Sandhini Agarwal, Katarina Slama, Alex Gray, John
  Schulman, Jacob Hilton, Fraser Kelton, Luke Miller, Maddie Simens, Amanda
  Askell, Peter Welinder, Paul Christiano, Jan Leike, and Ryan Lowe. 2022.
\newblock \href {https://openreview.net/forum?id=TG8KACxEON} {Training language
  models to follow instructions with human feedback}.
\newblock In \emph{Advances in Neural Information Processing Systems}.

\bibitem[{Patterson et~al.(2024)Patterson, Neumann, White, and
  White}]{patterson2024empiricaldesignreinforcementlearning}
Andrew Patterson, Samuel Neumann, Martha White, and Adam White. 2024.
\newblock \href {https://arxiv.org/abs/2304.01315} {Empirical design in
  reinforcement learning}.
\newblock \emph{Preprint}, arXiv:2304.01315.

\bibitem[{Rafailov et~al.(2023{\natexlab{a}})Rafailov, Sharma, Mitchell,
  Manning, Ermon, and Finn}]{rafailov2023direct}
Rafael Rafailov, Archit Sharma, Eric Mitchell, Christopher~D Manning, Stefano
  Ermon, and Chelsea Finn. 2023{\natexlab{a}}.
\newblock \href {https://openreview.net/forum?id=HPuSIXJaa9} {Direct preference
  optimization: Your language model is secretly a reward model}.
\newblock In \emph{Thirty-seventh Conference on Neural Information Processing
  Systems}.

\bibitem[{Rafailov et~al.(2023{\natexlab{b}})Rafailov, Sharma, Mitchell,
  Manning, Ermon, and Finn}]{NEURIPS2023_a85b405e}
Rafael Rafailov, Archit Sharma, Eric Mitchell, Christopher~D Manning, Stefano
  Ermon, and Chelsea Finn. 2023{\natexlab{b}}.
\newblock \href
  {https://proceedings.neurips.cc/paper_files/paper/2023/file/a85b405ed65c6477a4fe8302b5e06ce7-Paper-Conference.pdf}
  {Direct preference optimization: Your language model is secretly a reward
  model}.
\newblock In \emph{Advances in Neural Information Processing Systems},
  volume~36, pages 53728--53741. Curran Associates, Inc.

\bibitem[{Rasley et~al.(2020)Rasley, Rajbhandari, Ruwase, and
  He}]{DBLP:conf/kdd/RasleyRRH20}
Jeff Rasley, Samyam Rajbhandari, Olatunji Ruwase, and Yuxiong He. 2020.
\newblock \href {https://doi.org/10.1145/3394486.3406703} {Deepspeed: System
  optimizations enable training deep learning models with over 100 billion
  parameters}.
\newblock In \emph{KDD}, pages 3505--3506.

\bibitem[{Sakaguchi et~al.(2019)Sakaguchi, Bras, Bhagavatula, and
  Choi}]{sakaguchi2019winograndeadversarialwinogradschema}
Keisuke Sakaguchi, Ronan~Le Bras, Chandra Bhagavatula, and Yejin Choi. 2019.
\newblock \href {https://arxiv.org/abs/1907.10641} {Winogrande: An adversarial
  winograd schema challenge at scale}.
\newblock \emph{Preprint}, arXiv:1907.10641.

\bibitem[{Schulman et~al.(2018)Schulman, Moritz, Levine, Jordan, and
  Abbeel}]{schulman2018highdimensionalcontinuouscontrolusing}
John Schulman, Philipp Moritz, Sergey Levine, Michael Jordan, and Pieter
  Abbeel. 2018.
\newblock \href {https://arxiv.org/abs/1506.02438} {High-dimensional continuous
  control using generalized advantage estimation}.
\newblock \emph{Preprint}, arXiv:1506.02438.

\bibitem[{Schulman et~al.(2017)Schulman, Wolski, Dhariwal, Radford, and
  Klimov}]{schulman2017proximalpolicyoptimizationalgorithms}
John Schulman, Filip Wolski, Prafulla Dhariwal, Alec Radford, and Oleg Klimov.
  2017.
\newblock \href {https://arxiv.org/abs/1707.06347} {Proximal policy
  optimization algorithms}.
\newblock \emph{Preprint}, arXiv:1707.06347.

\bibitem[{Shao et~al.(2024)Shao, Wang, Zhu, Xu, Song, Bi, Zhang, Zhang, Li, Wu,
  and Guo}]{shao2024deepseekmathpushinglimitsmathematical}
Zhihong Shao, Peiyi Wang, Qihao Zhu, Runxin Xu, Junxiao Song, Xiao Bi, Haowei
  Zhang, Mingchuan Zhang, Y.~K. Li, Y.~Wu, and Daya Guo. 2024.
\newblock \href {https://arxiv.org/abs/2402.03300} {Deepseekmath: Pushing the
  limits of mathematical reasoning in open language models}.
\newblock \emph{Preprint}, arXiv:2402.03300.

\bibitem[{Talmor et~al.(2019)Talmor, Herzig, Lourie, and
  Berant}]{talmor-etal-2019-commonsenseqa}
Alon Talmor, Jonathan Herzig, Nicholas Lourie, and Jonathan Berant. 2019.
\newblock \href {https://doi.org/10.18653/v1/N19-1421} {{C}ommonsense{QA}: A
  question answering challenge targeting commonsense knowledge}.
\newblock In \emph{Proceedings of the 2019 Conference of the North {A}merican
  Chapter of the Association for Computational Linguistics: Human Language
  Technologies, Volume 1 (Long and Short Papers)}, pages 4149--4158,
  Minneapolis, Minnesota. Association for Computational Linguistics.

\bibitem[{Tang et~al.(2024)Tang, Guo, Zheng, Calandriello, Munos, Rowland,
  Richemond, Valko, {'A}vila Pires, and Piot}]{tang2024gpo}
Yunhao Tang, Zhaohan~Daniel Guo, Zeyu Zheng, Daniele Calandriello, R{'e}mi
  Munos, Mark Rowland, Pierre~Harvey Richemond, Michal Valko, Bernardo {'A}vila
  Pires, and Bilal Piot. 2024.
\newblock Generalized preference optimization: A unified approach to offline
  alignment.
\newblock \emph{arXiv preprint arXiv:2402.05749}.

\bibitem[{Wang et~al.(2023)Wang, Jiang, Yang, Liu, and Chen}]{wang2023fdpo}
Chaoqi Wang, Yibo Jiang, Chenghao Yang, Han Liu, and Yuxin Chen. 2023.
\newblock Beyond reverse kl: Generalizing direct preference optimization with
  diverse divergence constraints.
\newblock \emph{arXiv preprint arXiv:2309.16240}.

\bibitem[{Wang et~al.(2024)Wang, Paliotta, May, Rush, and
  Dao}]{junxiongdaniele2024mambainllama}
Junxiong Wang, Daniele Paliotta, Avner May, Alexander~M Rush, and Tri Dao.
  2024.
\newblock \href {https://openreview.net/forum?id=uAzhODjALU} {The mamba in the
  llama: Distilling and accelerating hybrid models}.
\newblock In \emph{The Thirty-eighth Annual Conference on Neural Information
  Processing Systems}.

\bibitem[{Xiao et~al.(2024)Xiao, Yuan, Zhu, Li, and Honavar}]{xiao2024caldpo}
Teng Xiao, Yige Yuan, Huaisheng Zhu, Mingxiao Li, and Vasant~G. Honavar. 2024.
\newblock Cal-dpo: Calibrated direct preference optimization for language model
  alignment.
\newblock \emph{arXiv preprint arXiv:2412.14516}.

\bibitem[{Yuan et~al.(2023)Yuan, Yuan, Tan, Wang, Huang, and
  Huang}]{yuan2023rrhf}
Zheng Yuan, Hongyi Yuan, Chuanqi Tan, Wei Wang, Songfang Huang, and Fei Huang.
  2023.
\newblock {RRHF}: Rank responses to align language models with human feedback
  without tears.
\newblock \emph{arXiv preprint arXiv:2304.05302}.

\bibitem[{Zellers et~al.(2019)Zellers, Holtzman, Bisk, Farhadi, and
  Choi}]{zellers2019hellaswagmachinereallyfinish}
Rowan Zellers, Ari Holtzman, Yonatan Bisk, Ali Farhadi, and Yejin Choi. 2019.
\newblock \href {https://arxiv.org/abs/1905.07830} {Hellaswag: Can a machine
  really finish your sentence?}
\newblock \emph{Preprint}, arXiv:1905.07830.

\end{thebibliography}

\appendix

\section{Methods}
\label{appendix:methods}

\subsection{RLHF \& DPO Background}
\label{appendix:methods_dpo}
In RLHF objective~\ref{eq:rlhf_objective}, access to the ground-truth target text $y$ may be limited or entirely unavailable. Instead, humans often provide preferences by indicating whether they prefer one output over another through a binary label. Furthermore, the reward function $r$ may be unknown beforehand, especially in preference-based optimization paradigms. These paradigms avoid explicitly modeling a reward function due to its computational costs \cite{NEURIPS2023_a85b405e}. Early implementations of RLHF addressed the aforementioned constrained optimization problem using PPO. PPO iteratively updates the policy $\pi_\theta$ by maximizing a clipped surrogate objective based on generalized advantage estimates \cite{schulman2017proximalpolicyoptimizationalgorithms, schulman2018highdimensionalcontinuouscontrolusing}. However, PPO requires maintaining and training a value network to estimate the advantage terms, which increases the computational overhead. PPO is also sensitive to the choice of hyperparameters \cite{engstrom2020implementationmattersdeeppolicy, ilyas2020closerlookdeeppolicy, patterson2024empiricaldesignreinforcementlearning}.

DPO \cite{rafailov2023direct} offers an alternative framework designed to align the language model $\pi_\theta$ directly with human preference data, without the need for explicit training of a reward function beforehand. This human preference data consists of triplets $(x, z_w, z_l)$, where $x$ is an input, $z_w$ is the preferred response, and $z_l$ is a less preferred response. DPO operates under the assumption that the probability of a specific preference can be accurately modeled by a Bradley-Terry model. Under this assumption, DPO optimizes the policy (i.e., the language model) by directly maximizing the log-likelihood of the preference data. This results in a simple classification objective based on the logistic function: $J_{\text{DPO}}(\theta) = $
\begin{equation}
\label{eq:dpo_objective}
\begin{aligned}
\log \sigma \left( \beta \log \frac{\pi_\theta(z_w|x)}{\pi_{\text{sft}}(z_w|x)} - \beta \log \frac{\pi_\theta(z_l|x)}{\pi_{\text{sft}}(z_l|x)} \right)
\end{aligned}
\end{equation}

The key advantage of DPO is that it circumvents the complexities associated with policy gradient methods by directly using preference information in the policy optimization process. This approach has resulted in increased training stability and improved computational efficiency, as it eliminates the need to maintain a separate value network, which is required by PPO. Moreover, empirical evaluations have suggested that PPO exhibits suboptimal performance compared to DPO \cite{rafailov2023direct}.

\subsection{Theorem~\ref{theory:implicit_po_mml}}
\label{appendix:mmpo_proof}
\begin{proof}
\label{appendix:proof:implicit_po_mml}

We begin by computing the derivative of the $\hat{J}^{rlhf}_{\text{MML}}(\theta)$ objective, which involves the `log-sum-exp' operation. Assuming that $s_l \le s_w$ for simplification, the derivative is:

\begin{multline*}
\nabla_{\theta} \hat{J}^{rlhf}_{\text{MML}}(\theta) = \nabla_{\theta} \log \sum_{i \in \{l, w\}} \exp \{s_{i}\} \\
= \nabla_{\theta} [s_{w} + \log \sum_{i \in \{l, w\}} \exp \{s_{i} - s_{w}\}] \\
= \nabla_{\theta} [s_{w} + \log \{1 + e^{(s_{l} - s_{w})}\}] \\
= \nabla_{\theta} s_{w} +  \frac{\nabla_{\theta} \{1 + e^{(s_{l} - s_{w})}\}}{1 + e^{(s_{l} - s_{w})}} \\
= \nabla_{\theta} s_{w} + \frac{e^{(s_{l} - s_{w})}}{1 + e^{(s_{l} - s_{w})}} \nabla_{\theta} (s_{l} - s_{w}) \\
= \frac{e^{(s_{l} - s_{w})}}{1 + e^{(s_{l} - s_{w})}} \nabla_{\theta} s_{l} +  \frac{1}{1 + e^{(s_{l} - s_{w})}} \nabla_{\theta} s_{w} \\
= \sigma(s_{l} - s_{w}) \nabla_{\theta} s_{l}  + \sigma(s_{w} - s_{l}) \nabla_{\theta} s_{w}
\end{multline*}

Given that $\nabla_{\theta} s_{l} = \nabla_{\theta} \log \pi_{\theta} (z_{l} | x)$ and $\nabla_{\theta} s_{w} = \nabla_{\theta} \log \pi_{\theta} (z_{w} | x)$, the proof is complete.
\end{proof}

\section{Experiments}

\subsection{Hyperparameters}
\label{appendix:experiments:hyper}
We tuned the hyperparameters for the preference optimization stage using a multi-stage grid search for each model size (135M, 360M, 1B, and 8B). In the first stage, we identified the most effective initial learning rate from the set $\{5 \times 10^{-4}, 10^{-4}, 5 \times 10^{-5}, 10^{-5}, 5 \times 10^{-6}, 10^{-6}\}$. This initial search involved training models for up to two epochs using default hyperparameter values: $\beta=0.1$ for all objectives, $\gamma/\beta=1.2$ for SimPO, and $r_\epsilon=0.6$ for MMPO. In the second stage, using the selected learning rate, we searched for an optimal $\beta$ value within the range $\{0.01, 0.05, 0.1, 0.5\}$. In the final stage, with the best-performing learning rate and $\beta$ fixed, we tuned $\gamma/\beta$ for SimPO over $\{0.3, 0.5, 1.0, 1.2, 1.4, 1.6\}$ and $r_\epsilon$ for MMPO over $\{0, 0.1, 0.2, 0.3, 0.4, 0.5, 0.6, 0.7, 0.8, 0.9, 1.0\}$.

To select the best-performing hyperparameters, we used the reward accuracy metric on the validation split of our preference datasets. This metric evaluates how frequently the internal score assigned to a chosen response is higher than the score for the corresponding rejected response. Specifically, the scoring functions were defined as follows: for DPO, we used $\beta \log \frac{\pi_\theta(z|x)}{\pi_{\text{SFT}}(z|x)}$; for SimPO, the score was $\beta \log \pi_\theta(z|x)$; and for MMPO, we used the scores $s_w$ and $s_l$ from Equation~\ref{eq:mml_gradient_score_def}, following the described in-batch reward normalization.

We observed that the reward accuracy metric consistently favored the smallest value, $\beta=0.01$, across all models. To provide a comprehensive analysis of the objectives under different $\beta$ settings, we subsequently trained the models for five epochs for each value of $\beta \in \{0.01, 0.05, 0.5\}$.

\begin{table}
\centering
\caption{The tuned hyperparameters across all models and objectives.}

\begin{tabular}{| c | c | c | c |}
\hline
Model-Objective & \multicolumn{3}{|c|}{Hyperparameter} \\
\hline
 & $lr$ & $\gamma/\beta$ & $r_{\epsilon}$  \\
\hline
SmolLM2-135M & & & \\
\small{MMPO} & 0.0005 & n/a & 0.9 \\
\small{DPO} & 0.0005  & n/a & n/a \\
\small{SimPO} & 0.0001 & 1.6 & n/a \\
\hline
SmolLM2-360M & & & \\
\small{MMPO} & 0.0005 & n/a & 0.9 \\
\small{DPO} & 0.00005 & n/a & n/a \\
\small{SimPO} & 0.0001 & 1.6 & n/a \\
\hline
Llama3.2-1B & & & \\
\small{MMPO} & 0.0005 & n/a & 0.9 \\
\small{DPO} & 0.0005 & n/a & n/a \\
\small{SimPO} & 0.0005 & 1.0 & n/a\\
\hline
Llama3-8B & & & \\
\small{MMPO} &  0.0001 & n/a & 0.9 \\
\small{DPO} & 0.00001 & n/a & n/a \\
\small{SimPO} & 0.0001 & 1.0 & n/a\\
\hline
\end{tabular}
\label{hyperparameters}
\end{table}

To train our models, we employed a cosine learning rate scheduler, which included a warmup phase constituting 10\% of the total training steps. This is a standard and effective technique in LLM optimization~\cite{loshchilov2017sgdrstochasticgradientdescent}. For all models, we configured a per-device batch size of 2 and used gradient accumulation over 16 steps. This resulted in a global batch size of 128. The training for all three preference optimization objectives (MMPO, DPO, and SimPO) was conducted on four L40 GPUs, leveraging the DeepSpeed Stage 2 optimizer for efficient multi-GPU training~\cite{DBLP:conf/kdd/RasleyRRH20}. We set the maximum prompt length to 1800 tokens and the maximum completion length to 512 tokens consistently across all models and objectives. For preference optimization with the Llama3-8B model, we used four H100 GPUs.

For the SmolLM2 models (135M and 360M parameters), we trained all parameters using the AdamW optimizer~\cite{loshchilov2019decoupledweightdecayregularization}. In contrast, for the Llama3.2-1B model, we utilized Low-Rank Adaptation (LoRA)~\cite{hu2021loralowrankadaptationlarge} on the attention and feedforward projection matrices. For the Llama3-8B model, we applied LoRA adaptation only on the attention projection matrices. For both Llama models, the LoRA configuration consisted of a rank ($r$) of 64, an alpha ($\alpha$) of 64, and a dropout rate of 0.01. The LoRA parameters were optimized using the 8-bit Adam optimizer to enhance memory efficiency~\cite{dettmers20228bitoptimizersblockwisequantization}.

Table~\ref{hyperparameters} summarizes the tuned hyperparameters across all the models and objectives.

\subsection{Entropy Gradient Approximation}
\label{appendix:experiments:ablation_entropy}

To compute the gradient of the entropy objective with respect to the parameters $\theta$, we use the following gradient approximation. We consider a mini-batch of $|B|$ text completions $z$. Each sample has a maximum sequence length of $T$ tokens. For simplicity, we consider padding all different sequences in the batch up to this maximum length $T$ while using a mask vector to cancel the effect of pad tokens.

\small
\begin{equation}
\begin{aligned}
\nabla_\theta\beta H\big(\pi_\theta(\cdot\mid x)\big) = \nabla_\theta P_{1} + \nabla_\theta P_{2}\\
\nabla_\theta P_{1} \approx \frac{\beta}{|B|}\sum^{|B|}_{i=1}\sum_{t=1}^T\nabla_\theta\log\pi_\theta(z^{(i)}_{<t}\mid x)\;H\!\big(\pi_\theta(\cdot\mid z^{(i)}_{<t}, x)\big)\\
\nabla_\theta P_{2} \approx \frac{\beta}{|B|}\sum^{|B|}_{i=1}\sum_{t=1}^{T}\nabla_\theta H\!\big(\pi_\theta(\cdot\mid z^{(i)}_{<t}, x)\big)
\end{aligned}
\end{equation}
\normalsize

\begin{proof}
\label{appendix:proof:entropy_proof}

We first show that the entropy objective can be rewritten in terms of the per-step entropies in the following equation:

\small
\begin{equation}
\label{eq:entropy_approximation}
\begin{aligned}
\beta H\big(\pi_\theta(\cdot\mid x)\big) &= -\beta\sum_{z}\pi_\theta(z\mid x)\log\pi_\theta(z\mid x)\\
&= -\beta\sum_{z}\pi_\theta(z\mid x)\sum_{t=1}^T\log\pi_\theta(z_t\mid z_{<t}, x)\\
&= -\beta\sum_{t=1}^T\sum_{z}\pi_\theta(z\mid x)\log\pi_\theta(z_t\mid z_{<t}, x)\\
&= -\beta\sum_{t=1}^T\sum_{z_{1:t}}\pi_\theta(z_{1:t}\mid x)\log\pi_\theta(z_t\mid z_{<t}, x)\\
&= \beta\sum_{t=1}^T\sum_{z_{<t}}\pi_\theta(z_{<t}\mid x)H\big(\pi_\theta(\cdot\mid z_{<t}, x)\big)
\end{aligned}
\end{equation}
\normalsize

We can have the following approximation for the gradient vector of the entropy with respect to the parameter set $\theta$:

\small
\begin{equation}
\begin{aligned}
\nabla_\theta\beta H\big(\pi_\theta(\cdot\mid x)\big) = \nabla_\theta P_{1} + \nabla_\theta P_{2}\\
\nabla_\theta P_{1} = \beta\sum_{t=1}^T\sum_{z_{<t}}\nabla_\theta \pi_\theta(z_{<t}\mid x) H\big(\pi_\theta(\cdot\mid z_{<t}, x)\big)\\
\nabla_\theta P_{2} = \beta\sum_{t=1}^T\sum_{z_{<t}}\pi_\theta(z_{<t}\mid x)\nabla_\theta H\big(\pi_\theta(\cdot\mid z_{<t}, x)\big)
\end{aligned}
\end{equation}
\normalsize

By using the score-function identity:

$\nabla_\theta\pi_\theta(z_{<t}\mid x)=\pi_\theta(z_{<t}\mid x) \nabla_\theta\log\pi_\theta(z_{<t}\mid x)$,
\normalsize we have $\nabla_\theta P_{1} = $
\small
\begin{equation}
\begin{aligned}
\beta\sum_{t=1}^T\sum_{z_{<t}}\pi_\theta(z_{<t}\mid x) \nabla_\theta\log\pi_\theta(z_{<t}\mid x) H \big(\pi_\theta(\cdot\mid z_{<t}, x)\big)
\end{aligned}
\end{equation}
\normalsize

By using the samples $z^{(i)}$ in the mini-batch to approximate the prefix sample expectations, we have our final approximation:

\small
\begin{equation}
\begin{aligned}
\nabla_\theta\beta H\big(\pi_\theta(\cdot\mid x)\big) = \nabla_\theta P_{1} + \nabla_\theta P_{2}\\
\nabla_\theta P_{1} \approx \frac{\beta}{|B|}\sum^{|B|}_{i=1}\sum_{t=1}^T\nabla_\theta\log\pi_\theta(z^{(i)}{<t}\mid x) H\big(\pi_\theta(\cdot\mid z^{(i)}{<t}, x)\big)\\
\nabla_{\theta} P_{2} \approx \frac{\beta}{|B|}\sum^{|B|}_{i=1}\sum_{t=1}^{T}\nabla_\theta H\big(\pi_\theta(\cdot\mid z^{(i)}_{<t}, x)\big)
\end{aligned}
\end{equation}
\normalsize

This finishes our proof.
\end{proof}

\end{document}